\documentclass[10pt,twocolumn,letterpaper]{article}
\usepackage{iccv}
\usepackage{times}
\usepackage{epsfig}
\usepackage{graphicx}
\usepackage{amsmath}
\usepackage{amssymb}

\usepackage{booktabs}
\usepackage{multirow}
\usepackage{verbatim}
\usepackage[ruled,linesnumbered]{algorithm2e}
\usepackage{subfigure}

\usepackage[pagebackref=true,breaklinks=true,letterpaper=true,colorlinks,bookmarks=false]{hyperref}

\iccvfinalcopy 

\def\iccvPaperID{3166} 

\ificcvfinal\fi

\begin{document}

\title{Enhancing Generalization of Universal Adversarial Perturbation through Gradient Aggregation}

\author{{Xuannan Liu, Yaoyao Zhong, Yuhang Zhang, Lixiong Qin, Weihong Deng\thanks{Corresponding author}}\\
Beijing University of Posts and Telecommunications\\
{\tt\small {\{liuxuannan, zhongyaoyao, zyhzyh, lxqin, whdeng\}@bupt.edu.cn}}
}

\maketitle
\ificcvfinal\thispagestyle{empty}\fi

\begin{abstract}
   Deep neural networks are vulnerable to universal adversarial perturbation (UAP), an instance-agnostic perturbation capable of fooling the target model for most samples. Compared to instance-specific adversarial examples, UAP is more challenging as it needs to generalize across various samples and models. In this paper, we examine the serious dilemma of UAP generation methods from a generalization perspective -- the gradient vanishing problem using small-batch stochastic gradient optimization and the local optima problem using large-batch optimization. To address these problems, we propose a simple and effective method called Stochastic Gradient Aggregation (SGA), which alleviates the gradient vanishing and escapes from poor local optima at the same time. Specifically, SGA employs the small-batch training to perform multiple iterations of inner pre-search. Then, all the inner gradients are aggregated as a one-step gradient estimation to enhance the gradient stability and reduce quantization errors. Extensive experiments on the standard ImageNet dataset demonstrate that our method significantly enhances the generalization ability of UAP and outperforms other state-of-the-art methods. The code is available at \url{https://github.com/liuxuannan/Stochastic-Gradient-Aggregation}.
\end{abstract}

\section{Introduction}
\label{sec:intro}
Deep neural networks (DNNs) have achieved significant success in computer vision~\cite{hinton2012imagenet,szegedy2015going,simonyan2014very,he2016deep,howard2017mobilenets,hu2018squeeze}, but are widely known to be vulnerable to adversarial examples~\cite{szegedy2013intriguing,moosavi2016deepfool,papernot2016limitations,goodfellow2014explaining,kurakin2018adversarial,madry2017towards,xiao2018generating}. A more critical property of adversarial examples is that they have shown good transferability between different models~\cite{liu2016delving,papernot2017practical,dong2018boosting,dong2019evading,lin2019nesterov,xie2019improving,zhong2020towards,wang2021enhancing,qin2022boosting}. Unlike the instance-specific adversarial examples, a recent work~\cite{moosavi2017universal} reveals the existence of universal adversarial perturbation (UAP) which can deceive the majority of samples. 

\begin{figure}[!t]
	\centering
	\includegraphics[width=0.90\linewidth]{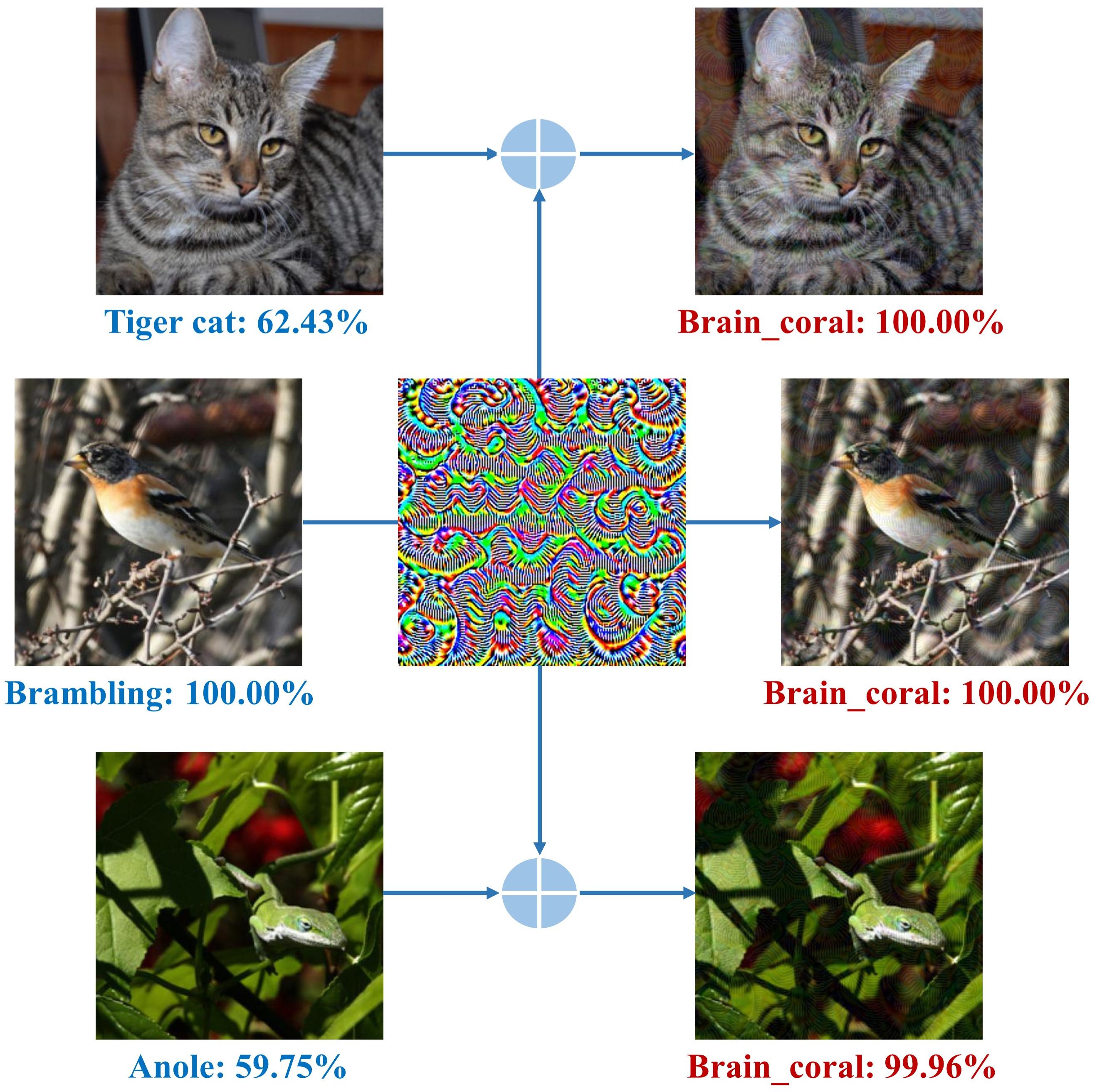}
	\caption{A universal adversarial perturbation is applied to images belonging to different categories to get visually similar adversarial examples with high attack success rates in the white-box scenario. \textbf{Left images:} the original natural images. \textbf{Central image:} the UAP by applying SGA on the VGG16 model (rescaled to [0,255]). \textbf{Right images:} the adversarial images.}
	\label{uap_img}
\end{figure}

Compared to the instance-specific adversarial examples, the generation of UAP is more challenging. Since most UAPs are generated with limited training samples and are expected to be applied to various unknown samples~\cite{khrulkov2018art,poursaeed2018generative,shafahi2020universal,li2022learning} and even a variety of tasks~\cite{xie2021enabling,ding2021beyond,rampini2021universal,berger2022stereoscopic,peng2022fingerprinting,zhong2022opom}, the diversity of samples and models has created dual demands for higher generalization ability of UAPs. Therefore, we aim to investigate the limitation of the current UAPs and improve the generalization ability.

\begin{figure}[!t]
	\centering
	\includegraphics[width=0.90\linewidth]{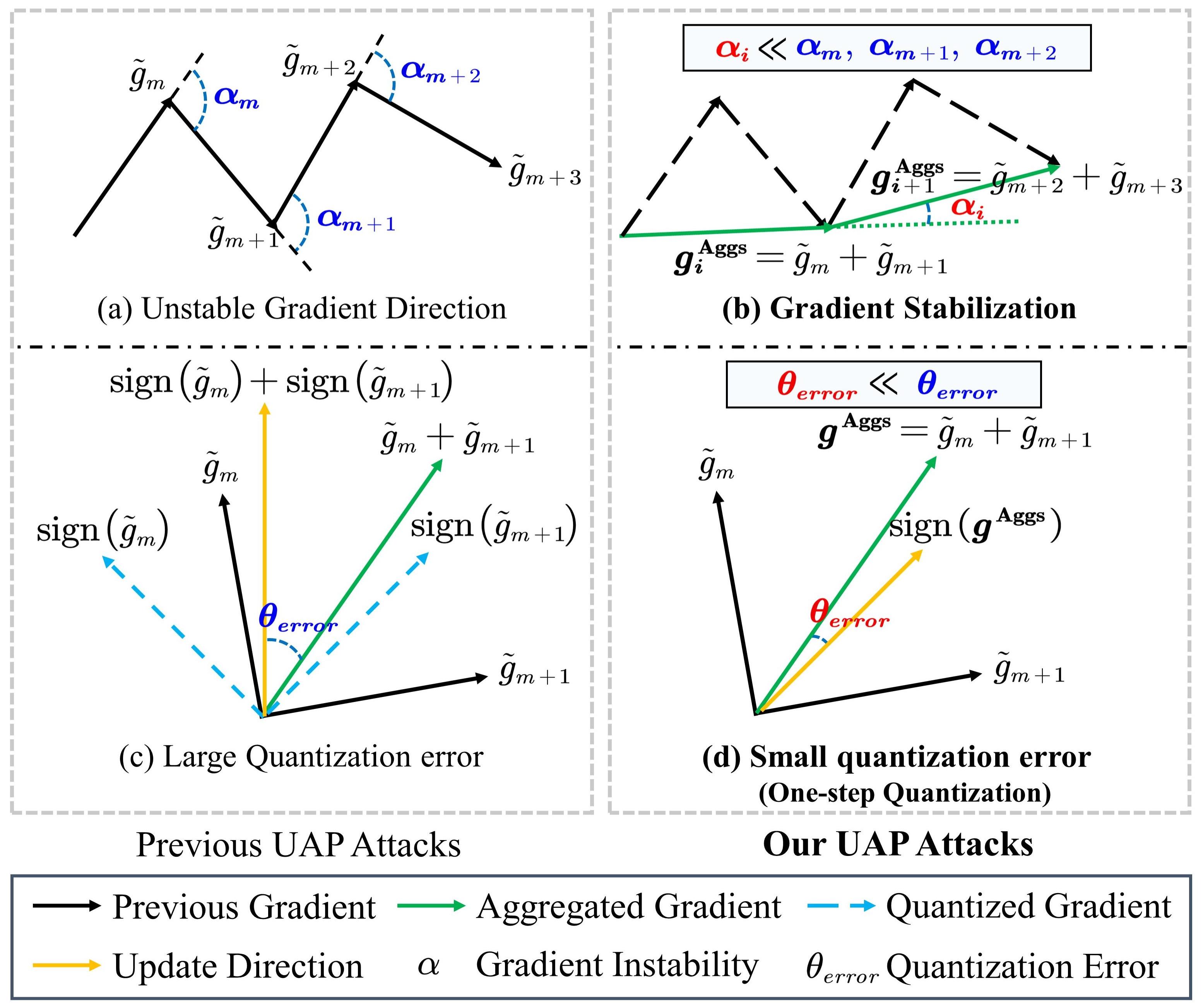}
	\caption{Illustration of the two issues, \textit{i.e.}, gradient instability and quantization error, under previous UAP attacks and our attacks, respectively. Where $\tilde{g}_m$ and $g^{\mathrm{Aggs}}$ denote the pre-search gradient and the aggregated gradient respectively. In (b), we use the subscript $i$ to distinguish between different $g_{i}^{\mathrm{Aggs}}$ in the outer iteration. (a) and (c): Previous works sequentially quantize the unstable gradients to accumulate the adversarial perturbations, resulting in severe gradient deviation. (b) and (d): Our work accumulates the pre-search gradients as a one-step gradient update with low variance before quantization to reduce the gradient deviation.}
	\label{gradient_vanish}
\end{figure}

Despite many works on UAP, there are two main issues in the generation of UAP. (1) Gradient instability. Due to the inherent difference of samples and model parameters, the optimization paths for adversarial perturbations vary widely, as shown in Fig \ref{gradient_vanish} (a). Such instability may hinder the optimization of adversarial attacks in the correct direction~\cite{wang2021enhancing,xiong2022stochastic,wang2022boosting}. (2) Quantization error. The adversarial attack methods generally exploit the sign operations for quickly crafting adversarial perturbations according to the linearity of models. Frequent use of sign will accumulate a large amount of quantization errors~\cite{cheng2021fast,zhang2022revisiting} in Fig \ref{gradient_vanish} (c). Furthermore, when quantizing gradients with high fluctuations using the sign operations, the large values of the gradients in the forward optimization process can be easily eliminated by the negative small gradient values in the backward process, leading to the gradient vanishing phenomenon.

The generation of UAP can be described as a non-convex optimization problem and adopt stochastic gradient descent (SGD) with mini-batch training. Therefore, an intuitive idea is to use large-batch methods to stabilize gradient update directions, thereby suppressing gradient vanishing. However, when optimizing via SGD, small-batch methods have been shown to converge more easily to flat local minima than large-batch methods, thus effectively reducing the generalization gap~\cite{keskar2016large}. Inevitably, the generalization of UAP faces the grim dilemma of choosing small-batch training with the gradient vanishing problem or choosing large-batch training with the local optima problem.

In this paper, we propose a novel method called stochastic gradient aggregation (SGA), to address the gradient vanishing and the local optima problems to enhance the generalization of UAP. Specifically, our method consists of the inner-outer iterations. At each outer iteration for the gradient estimating, we randomly select multiple small-batch samples to perform the inner iterations for pre-search. Then we aggregate all inner gradients as a one-step iterative gradient for updating UAP, as illustrated in Fig \ref{gradient_vanish} (b) and (d). The key idea is to cope with gradient vanishing by enhancing the gradient stability and decreasing the use of quantization operations while introducing the noisy gradients to escape from sharp local optima. To the best of our knowledge, this is the first work to investigate the limitation of existing universal attacks through the perspective of generalization. The main contributions of our paper are as follows:
\begin{itemize}
	\item We investigate two issues behind the low generalization ability of existing UAP works, \textit{i.e.}, gradient instability and quantization error, and further identify the gradient vanishing phenomenon when the iterative gradients have high fluctuations.
	\item We propose stochasitc gradient aggregation (SGA) that stabilizes the update directions and reduces quantization errors to alleviate the gradient vanishing in the small-batch optimization.
	\item Our method can be easily integrated with transferability attack methods. Extensive experiments demonstrate the superior generalization of UAP generated by the proposed SGA  compared to the state-of-the-art methods under various attack settings.
\end{itemize}

\section{Related work}
\label{sec:related work}
\textbf{Instance-specific attack methods.} Szegedy \etal~\cite{szegedy2013intriguing} first found the adversarial example and proposed an attack method based on box constraints, named as L-BFGS. By exploiting the decision boundary, Moosavi-Dezfooli \etal~\cite{moosavi2016deepfool} proposed to generate the minimal perturbation at each step, called DeepFool. Moreover, Papernot \etal~\cite{papernot2016limitations} constructed adversarial saliency maps to search for critical input regions for generating adversarial examples. To improve efficiency, Goodfellow \etal~\cite{goodfellow2014explaining} performed a one-step attack updating along the sign direction of the gradient, widely known as the fast gradient sign method (FGSM). Kurakin \etal~\cite{kurakin2018adversarial} further extended it to an iterative version of the FGSM, called I-FGSM. In addition, projected gradient descent (PGD)~\cite{madry2017towards} is another gradient-based iterative attack and has demonstrated excellent attack performance. To ensure the naturalness of adversarial examples, Xiao \etal~\cite{xiao2018generating} proposed to exploit Generative Adversarial Networks to directly synthesize tiny perturbations.

\textbf{Universal attack methods.} Previous works on UAP mainly focus on the cross-sample university. Moosavi-Dezfooli \etal~\cite{moosavi2017universal} first revealed the existence of UAPs and generated them by iteratively aggregating perturbation vectors obtained by the DeepFool method. SV-UAP~\cite{khrulkov2018art} exploited a few feature maps to calculate the singular vectors of the Jacobian matrices for crafting UAP. Poursaeed \etal~\cite{poursaeed2018generative} and Mopuri \etal~\cite{mopuri2018nag} both proposed to use generative models to synthesize UAP, named GAP and NAG respectively. To make UAP only misclassify the target class, Zhang \etal~\cite{zhang2020cd} proposed a class discriminative UAP (CD-UAP). Other works of Zhang~\cite{zhang2020understanding,zhang2021data} utilized dominant features and the cosine similarity loss to generate UAP, denoted as DF-UAP and Cos-UAP respectively. Besides, Shafahi \etal~\cite{shafahi2020universal}, Matachana \etal~\cite{matachana2020robustness} and Co \etal~\cite{co2021universal} all introduced stochastic gradient method for efficiently generating UAP by using PGD with mini-batch training. Recently, Li \etal~\cite{li2022learning} proposed to integrate instance-specific and universal attacks from a feature perspective to generate a more powerful UAP, called AT-UAP. However, unlike research~\cite{keskar2016large,hoffer2017train,masters2018revisiting,zhang2021relative,zhang2023unsupervised} on the generalization of model training, there are few works on the generalization of UAP.

\textbf{Gradient optimization.} Various of gradient optimization methods are proposed to improve the model transferability of adversarial examples. Dong \etal~\cite{dong2018boosting} proposed to add the momentum item to stabilize update directions and escape from poor local maxima. Furthermore, Lin \etal~\cite{lin2019nesterov} introduced Nesterov accelerated gradient and scale-invariant property. Wang~\cite{wang2021enhancing} proposed to shrink the gradient variance at each step to converge to better local optima. To encourage adversarial examples to converge flat local minimums, Qin \etal~\cite{qin2022boosting} proposed a new attack method by additionally injecting worst-case perturbations at each step to avoid overfitting to the surrogate model. 
\section{Methodology}
\subsection{Preliminaries of UAP}
\textbf{Problem formulation.} The objective of universal adversarial attack is to craft a single perturbation $\delta$ to fool the target model $f$ for most sample $x_i\in X$. In this way, we can consider the following optimization problem,
\begin{equation}
	\begin{aligned}
		\underset{\delta}{\mathrm{arg}\max}\frac{1}{n}&\sum_{i=1}^n{\mathcal{L} \left( f\left( x_i+\delta \right) ,y_i \right)},\\
		&s.t. \left\| \delta \right\| _{\infty}\leqslant \epsilon,
	\end{aligned}
\end{equation}
where $f\left( x \right)$ represents the output of the model $f$ with the input $x$, $y_i$ is the corresponding label of the input $x_i$ calculated by $y_i=\mathrm{arg}\max f\left( x_i \right)$. $\mathcal{L} \left( \cdot \right)$ denotes the adversarial loss used for generating adversarial perturbations. $\epsilon$ limits the maximum deviation of UAP. 

\textbf{SPGD.} Among existing adversarial attack methods, PGD~\cite{madry2017towards,shafahi2019adversarial} is considered to be a strong and efficient method. To accelerate the generation of UAP, Shafihi \etal~\cite{shafahi2020universal} first combined the stochastic gradient method with the PGD attack method to solve the above optimization problem, termed SPGD by:
\begin{equation}
	\tilde{g}_k=\frac{1}{\left| B_k \right|}\sum_{x_i\in B_k}{\nabla _{\delta}\mathcal{L} \left( x_i+\delta _k \right)} , 
\end{equation}
\begin{equation}
	\delta _{k+1}=\delta _k+\alpha \cdot \mathrm{sign}\left( \tilde{g}_k \right) ,
\end{equation}
where $B_k$ is a batch of samples in the data set, $\alpha$ is a step size and $\mathrm{sign}\left( \cdot \right)$ denotes the sign function.

\textbf{Loss function.} Two widely used adversarial loss functions are the cross-entropy (CE) loss~\cite{xie2019improving} and the logit loss~\cite{zhao2021success}, which will be both implemented in later experiments. To prevent the infinite cross-entropy loss value of a single sample from dominating the optimization of the entire objective function, we use the clipped version ~\cite{shafahi2020universal} of cross-entropy loss $\tilde{\mathcal{L}}_{\mathrm{CE}}=\min \left\{ \mathcal{L} _{\mathrm{CE}},\beta \right\}$, where $\beta$ represents the threshold for cross-entropy loss.

\subsection{Stochastic gradient aggregation}
\subsubsection{Gradient vanishing}
\label{gradient vanishing}
In UAP attacks, gradient instability is a normal issue due to the sample diversity and sign is a regular operation for quickly crafting adversarial examples. However, we discover that the sign operations will lead to huge optimization errors when the iterative gradients are strongly unstable. To illustrate this phenomenon, consider a simple toy example. Let $\tilde{g}_m$ and $\tilde{g}_{m+1}$ be gradients at iteration $m$ and $m+1$:
\begin{equation}
	\tilde{g}_m=\left[ \cdots ,\underline{-0.01},0.10,0.05,\underline{0.70},\cdots \right] ^{\mathrm{T}} ,
\end{equation}
\begin{equation}
	\tilde{g}_{m+1}=\left[ \cdots ,\underline{1.00},0.02,0.30,\underline{-0.01},\cdots \right] ^T .
\end{equation}
When using sign on the gradients for accumulating perturbation $\delta$, large value for the right region of $\tilde{g}_m$ is eliminated by the small negative value of $\tilde{g}_{m+1}$. Similarly in the left region, the sign operations lead to a significant deviation in the current optimization direction:
\begin{equation}
	\mathrm{sign}\left( \tilde{g}_m \right) =\left[ \cdots ,\underline{-1},1,1,\underline{1},\cdots \right] ^T,
\end{equation}
\begin{equation}
	\mathrm{sign}\left( \tilde{g}_{m+1} \right) =\left[ \cdots ,\underline{1},1,1,\underline{-1},\cdots \right] ^T,
\end{equation}
\begin{equation}
	\begin{aligned}
	\delta &=\alpha \cdot \mathrm{sign}\left( \tilde{g}_m \right) +\alpha \cdot \mathrm{sign}\left( \tilde{g}_{m+1} \right) \\
	&=\alpha \cdot \left[ \cdots ,\underline{0},2,2,\underline{0},\cdots \right] ^T .
	\end{aligned}
\end{equation}
Here we refer to this phenomenon as gradient vanishing, which is caused by the combination of gradient instability and sign operations.

\begin{figure*}[ht!]
	\centering
	\includegraphics[width=0.95\linewidth]{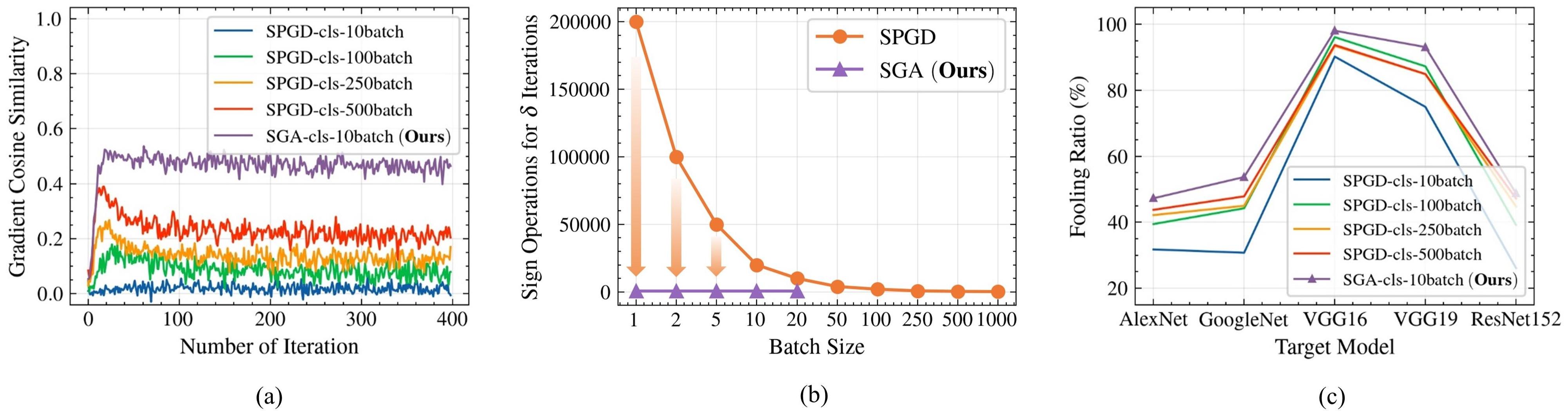}
	\caption{Analysis of two issues, \textit{i.e.}, gradient instability and quantization error under different batch sizes in stochastic gradient universal attacks. (a) represents the gradient cosine similarity between iterations; (b) represents the number of times the sign operator is used for the updates of UAP; (c) represents the average fooling ratio of five models. The gradient cosine similarity is used to describe the gradient stability and is obtained by calculating the cosine value between the forward gradient and the backward gradient, with a larger cosine similarity indicating a more stable gradient update direction.}
	\label{gradient_cos}
\end{figure*}

\subsubsection{Large-batch or small-batch}
Since the optimization problem for the generation of UAP can be solved by stochastic gradient algorithm with mini-batch training, it is easy to find using large-batch methods can effectively cope with gradient vanishing. However, in the deep learning tasks, recent works~\cite{keskar2016large,hoffer2017train,masters2018revisiting} verify that the large-batch methods tend to be attracted to sharp minima leading to the large generalization gap. Similarly, when UAP locates at a sharp local extremum in large-batch optimization, the difference in samples and models will result in a significant change in attack loss, making the UAP unable to generalize to unknown samples and models. The small-batch optimization has the advantage of escaping from poor local optima by exploiting the inherent noise in the gradient estimation. However, as shown in Fig \ref{gradient_vanish} (a) and (c), we empirically find that, it will further deteriorate two issues, \textit{i.e.}, gradient instability and quantization error, which will cause severe gradient vanishing, as illustrated in Section \ref{gradient vanishing}.

To further illustrate the impact of gradient instability and quantization error, we explore the relationship between two issues and attack performance under different batch sizes. First, in Fig \ref{gradient_cos} (a), the stability between the gradients obtained by small-batch training in the regular UAP attacks is poor. The large-batch methods can effectively stable the gradient update directions. Second, Fig \ref{gradient_cos} (b) illustrates that the sign operators are much more frequently used for the UAP iteration in small-batch methods than in large-batch methods, indicating that it is easier to accumulate larger quantization errors. Furthermore, Fig \ref{gradient_cos} (c) suggests that by enhancing gradient stability and reducing the usage of sign, large-batch methods can mitigate the gradient vanishing to improve the attack performance, but optimization with over large batch size also inhibits the generalization of UAP. In this way, both the gradient vanishing in small-batch training and the optimization problem in large-batch training will make the generalization of UAP in trouble.

\begin{algorithm}
	\caption{The SGA attack algorithm}\label{algorithm}
	\SetKwData{Left}{left}\SetKwData{This}{this}\SetKwData{Up}{up}
	\SetKwInOut{Input}{input}\SetKwInOut{Output}{output}
	\Input{A surrogate model $f$, loss function $\mathcal{L}$}
	\Input{The training image set $\boldsymbol{X}$, large-batch $\boldsymbol{x}^{\mathrm{LB}}$, small-batch $\boldsymbol{x}^{\mathrm{SB}}$}
	\Input{Maximum perturbation magnitude $\epsilon$, number of epochs $T$, step size $\alpha$}
	\Output{A universal adversarial perturbation $\delta$}
	Initialize $\delta=0$\;
	\For{$t=0\,\,to\,\,T-1$}{
		\For{$\boldsymbol{x}^{\mathrm{LB}}\in \boldsymbol{X}$}{
			$\delta _{0}^{\mathrm{inner}}=\delta$\;
			$g^{\mathrm{Aggs}}=0$\;
			\For{$m=0\,\,to\,\,M-1$}{
				Random select $\boldsymbol{x}_{m}^{\mathrm{SB}}\in \boldsymbol{x}^{\mathrm{LB}}$\;
				$\tilde{g}_m=\frac{1}{\left| \boldsymbol{x}^{\mathrm{SB}} \right|}\nabla _{\delta}\mathcal{L} \left( \boldsymbol{x}_{m}^{\mathrm{SB}}+\delta _{m}^{\mathrm{inner}} \right)$\;												
				$\delta _{m+1}^{\mathrm{inner}}=\mathrm{Clip}_{\delta}^{\epsilon}\left( \delta _{m}^{\mathrm{inner}}+\alpha \cdot \mathrm{sign}\left( \tilde{g}_m \right) \right)$\;
				$g^{\mathrm{Aggs}}\gets g^{\mathrm{Aggs}}+\tilde{g}_m$\;}
			$\delta \gets \mathrm{Clip}_{\delta}^{\epsilon}\left( \delta +\alpha \cdot \mathrm{sign}\left( g^{\mathrm{Aggs}} \right) \right)$\;}
	}
	\textbf{return} $\delta$.
\end{algorithm}

\subsubsection{Attack algorithms}
As discussed above, our target is to enhance gradient stability and reduce quantization errors when optimizing with small-batch methods. 

In this paper, instead of using the average gradient of a batch sample in conventional UAP attacks, a simple strategy is to aggregate multi-step noisy forward gradients for a one-step quantization update at each iteration. Specifically, we first randomly select multiple small-batch samples $\boldsymbol{x}_{m}^{\mathrm{SB}}$ from a large-batch set $\boldsymbol{x}^{\mathrm{LB}}$ to perform pre-search by updating the inner adversarial perturbation $\delta _{m}^{\mathrm{inner}}$:
\begin{equation}
	\tilde{g}_m=\frac{1}{\left| \boldsymbol{x}^{\mathrm{SB}} \right|}\nabla _{\delta}\mathcal{L} \left( \boldsymbol{x}_{m}^{\mathrm{SB}}+\delta _{m}^{\mathrm{inner}} \right) , 
\end{equation}
\begin{equation}
	\delta _{m+1}^{\mathrm{inner}}=\mathrm{Clip}_{\delta}^{\epsilon}\left( \delta _{m}^{\mathrm{inner}}+\alpha \cdot \mathrm{sign}\left( \tilde{g}_m \right) \right) . 
\end{equation}
Then, we accumulate the stochastic gradients $\tilde{g}_m$ for updating the aggregated gradient $g^{\mathrm{Aggs}}$:
\begin{equation}
	g^{\mathrm{Aggs}}\gets g^{\mathrm{Aggs}}+\tilde{g}_m .
\end{equation}
After aggregating all the inner gradients into a one-step gradient of the outer iteration, we update the outer adversarial perturbation $\delta$ using the aggregated gradient:
\begin{equation}
	\delta \gets \mathrm{Clip}_{\delta}^{\epsilon}\left( \delta +\alpha \cdot \mathrm{sign}\left( g^{\mathrm{Aggs}} \right) \right) ,
\end{equation}
where $\alpha$ is a step size and Clip$\left( \cdot \right)$ operation constrains the perturbation amplitude under the $l_{\infty}$ norm. 

By means of accumulating more important gradient information, the iterative gradient estimate items are more accurate with a small variance while greatly decreasing quantization operations, thus suppressing gradient vanishing. Moreover, the existence of noise gradients can help escape from poor local optima. The algorithm of stochastic gradient aggregation is summarized in the Algorithm \ref{algorithm}.

In short, our method differs from SPGD in that SGA has inner iterations, where SGA obtains multiple noisy gradients through $M$ updates. Model transferability methods in instance-specific adversarial attacks (\eg Momentum~\cite{dong2018boosting}, Nesterov accelerated gradient~\cite{lin2019nesterov}, \textit{etc.}) can be easily integrated with SGA in the inner iteration.

\begin{table}[!h]
	\begin{center}
		\renewcommand{\arraystretch}{1.2}
		\resizebox{\linewidth}{!}{
			\begin{tabular}{|c|cccccc|}
				\hline
				Method        & AlexNet        & GoogleNet      & VGG16          & VGG19          & ResNet152      & Average        \\ \hline \hline
				UAP~\cite{moosavi2017universal}           & 93.30           & 78.90           & 78.30           & 77.80           & 84.00           & 82.46          \\
				SV-UAP~\cite{khrulkov2018art}        & -              & -              & 52.00           & 60.00           & -              & 56.00             \\
				NAG~\cite{mopuri2018nag}           & 96.44          & 90.37          & 77.57          & 83.78          & 87.24          & 87.08          \\
				GAP~\cite{poursaeed2018generative}           & -              & 82.70           & 83.70           & 80.10           & -              & 82.17          \\
				DF-UAP~\cite{zhang2020understanding}        & 96.17          & 88.94          & 94.30          & 94.98          & 90.08          & 92.89          \\
				Cos-UAP~\cite{zhang2021data}       & 96.50           & 90.50           & 97.40           & 96.40           & 90.20           & 94.20           \\
				AT-UAP~\cite{li2022learning}        & 97.01          & 90.82          & 97.51          & 97.56          & 91.52          & 94.88          \\
				\textbf{Ours} & \textbf{97.43} & \textbf{92.12} & \textbf{98.36} & \textbf{97.69} & \textbf{94.04} & \textbf{95.93} \\ \hline
			\end{tabular}	
		}
	\end{center}
	\caption{The fooling ratio (\%) in the white-box setting by various UAP attack methods. The UAPs are crafted on five normally trained models, \textit{i.e.}, AlexNet, GoogleNet, VGG16, VGG19, and ResNet152.}
	\label{white-box generalization}
\end{table}

\begin{table*}[!h]
	\begin{center}
		\renewcommand{\arraystretch}{1.0}
		\scalebox{0.870}{
			\begin{tabular}{|c|c|cccccc|}
				\hline
				Model                                            & Method     & AlexNet         & GoogleNet       & VGG16           & VGG19           & ResNet152       & Average        \\ \hline \hline
				\multirow{6}{*}{AlexNet}                         & UAP        & 86.53*          & 27.82           & 37.67           & 35.47           & 20.99           & 41.70          \\
				& GAP        & 89.06*          & 33.05           & 52.02           & 48.60           & 28.70           & 50.29          \\ \cline{2-8} 
				& SPGD-logit & 95.23*          & 37.68           & 57.62           & 53.86           & 30.39           & 54.96          \\
				& SGA-logit (\textbf{Ours})  & \textbf{96.60*} & \textbf{46.18}  & \textbf{63.82}  & \textbf{59.52}  & \textbf{34.95}  & \textbf{60.21} \\ \cline{2-8} 
				& SPGD-cls   & 95.97*          & 42.78           & 58.59           & 54.03           & 30.52           & 56.38          \\
				& SGA-cls (\textbf{Ours})    & \textbf{97.23*} & \textbf{48.97}  & \textbf{66.46}  & \textbf{60.60}  & \textbf{35.29}  & \textbf{61.71} \\ \hline
				\multirow{6}{*}{GoogleNet}                       & UAP        & 44.63           & 74.51*          & 52.92           & 52.63           & 32.74           & 51.48          \\
				& GAP        & 56.55           & 76.60*          & 69.65           & 67.59           & 46.29           & 63.34          \\ \cline{2-8} 
				& SPGD-logit & 53.73           & \textbf{88.88*} & 77.13           & 74.88           & 54.21           & 69.77          \\
				& SGA-logit (\textbf{Ours})  & \textbf{65.92}  & 86.54*          & \textbf{80.85}  & \textbf{78.18}  & \textbf{56.53}  & \textbf{73.60} \\ \cline{2-8} 
				& SPGD-cls   & 50.10           & \textbf{90.70*} & 76.47           & 73.88           & 51.65           & 68.56          \\
				& SGA-cls (\textbf{Ours})    & \textbf{61.34}  & 90.64*          & \textbf{81.11}  & \textbf{79.06}  & \textbf{55.69}  & \textbf{73.57} \\ \hline
				\multirow{6}{*}{VGG16}                           & UAP        & 33.35           & 33.51           & 76.73*          & 64.14           & 29.39           & 47.42          \\
				& GAP        & 22.33           & 42.50           & 82.21*          & 76.30           & 29.46           & 50.56          \\ \cline{2-8} 
				& SPGD-logit & 39.20           & 54.12           & 93.56*          & 86.78           & 48.78           & 64.49          \\
				& SGA-logit (\textbf{Ours})  & \textbf{42.74}  & \textbf{57.58}  & \textbf{95.01*} & \textbf{89.99}  & \textbf{52.25}  & \textbf{67.51} \\ \cline{2-8} 
				& SPGD-cls   & 42.17           & 44.97           & 93.45*          & 84.89           & 44.77           & 62.05          \\
				& SGA-cls (\textbf{Ours})    & \textbf{47.26}  & \textbf{53.70}  & \textbf{98.04*} & \textbf{93.08}  & \textbf{48.78}  & \textbf{68.17} \\ \hline
				\multirow{6}{*}{VGG19}                           & UAP        & 34.45           & 35.21           & 65.46           & 77.79*          & 28.49           & 48.28          \\
				& GAP        & \textbf{52.71}  & 49.11           & 75.08           & 79.11*          & 35.21           & 58.24          \\ \cline{2-8} 
				& SPGD-logit & 44.47           & 54.94           & 86.63           & 93.87*          & 48.15           & 65.61          \\
				& SGA-logit (\textbf{Ours})  & \textbf{47.20}  & \textbf{57.53}  & \textbf{91.33}  & \textbf{96.07*} & \textbf{50.13}  & \textbf{68.45} \\ \cline{2-8} 
				& SPGD-cls   & 43.15           & 47.51           & 84.62           & 92.94*          & 42.09           & 62.06          \\
				& SGA-cls (\textbf{Ours})    & \textbf{46.53}  & \textbf{51.53}  & \textbf{93.35}  & \textbf{97.39*} & \textbf{45.47}  & \textbf{66.86} \\ \hline
				\multicolumn{1}{|l|}{\multirow{6}{*}{ResNet152}} & UAP        & 40.59           & 44.89           & 64.00           & 60.50           & 79.39*          & 57.87          \\
				\multicolumn{1}{|l|}{}                           & GAP        & 47.74           & 52.43           & 64.41           & 63.41           & 67.76*          & 59.15          \\ \cline{2-8} 
				\multicolumn{1}{|l|}{}                           & SPGD-logit & 43.02           & 54.13           & 76.47           & 74.40           & 90.83*          & 67.77          \\
				\multicolumn{1}{|l|}{}                           & SGA-logit (\textbf{Ours})  & \textbf{47.07}  & \textbf{56.91}  & \textbf{78.84}  & \textbf{77.86}  & \textbf{93.27*} & \textbf{70.79} \\ \cline{2-8} 
				\multicolumn{1}{|l|}{}                           & SPGD-cls   & 46.48           & 53.86           & 76.12           & 73.15           & 90.75*          & 68.07          \\
				\multicolumn{1}{|l|}{}                           & SGA-cls (\textbf{Ours})    & \textbf{50.12}  & \textbf{61.61}  & \textbf{80.98}  & \textbf{78.29}  & \textbf{93.16*} & \textbf{72.83} \\ \hline
			\end{tabular}
		}
	\end{center}
	\caption{The fooling ratio (\%) on five models in the black-box setting by regular UAP attack methods. The UAPs are crafted on AlexNet, GoogleNet, VGG16, VGG19, and ResNet152 respectively. * indicates the white-box model.}
	\label{base}
\end{table*} 

\begin{table*}[!ht]
	\begin{center}
	\renewcommand{\arraystretch}{1.0}
	\scalebox{0.870}{
		\begin{tabular}{|c|c|cccccc|}
			\hline
			Method & \multicolumn{1}{l|}{\# samples} & AlexNet        & GoogleNet      & VGG16          & VGG19          & ResNet152      & \multicolumn{1}{l|}{Average} \\ \hline \hline
			UAP    & 500                             & 57.33          & 16.61          & 25.29          & 25.04          & 19.11          & 28.68                        \\
			GAP    & 500                             & 86.89          & 57.07          & 70.40          & 65.89          & 47.58          & 65.57                        \\
			SPGD   & 500                             & 92.35          & 41.68          & 81.70          & 75.74          & 23.44          & 62.98                        \\
			SGA (\textbf{Ours})    & 500                             & \textbf{94.03} & \textbf{68.33} & \textbf{89.83} & \textbf{88.70} & \textbf{52.12} & \textbf{78.60}               \\ \hline
		\end{tabular}
	}
	\end{center}
	\caption{The fooling ratio (\%) on five models in the limit-sample setting by regular UAP attack methods. The UAPs are crafted on AlexNet, GoogleNet, VGG16, VGG19, and ResNet152 respectively.}
	\label{limit_500}
\end{table*} 

\section{Experiment}
\subsection{Experimental setup}
\textbf{Dataset.} Following~\cite{moosavi2017universal}, we randomly choose 10 images from each category in the ImageNet training set~\cite{russakovsky2015imagenet}, a total of 10,000 images, for the generation of UAP. Then we evaluate our method on the ImageNet validation set which contains 50,000 images. 

\textbf{Models.} We use five normally trained models including AlexNet~\cite{hinton2012imagenet}, Googlenet~\cite{szegedy2015going}, VGG-16~\cite{simonyan2014very}, VGG-19~\cite{simonyan2014very}, and ResNet152~\cite{he2016deep}. 

\textbf{Evaluation metrics.} To effectively evaluate the attack performance of our method, we report the fooling ratio metric which is most widely implemented in the UAP tasks~\cite{moosavi2017universal,poursaeed2018generative,mopuri2018nag,zhang2020understanding,zhang2021data,li2022learning}. The fooling ratio is obtained by calculating the proportion of samples with labels changes when applying UAP. 

\textbf{Baselines.} The proposed method is compared with the following UAP methods in the white-box attack scenario: UAP~\cite{moosavi2017universal}, SV-UAP~\cite{khrulkov2018art}, NAG~\cite{mopuri2018nag}, GAP~\cite{poursaeed2018generative}, DF-UAP~\cite{zhang2020understanding}, Cos-UAP~\cite{zhang2021data} and AT-UAP~\cite{li2022learning}. In the black-box attack setting, we regard SPGD~\cite{shafahi2020universal} as our baseline, and also consider other regular methods, \textit{i.e.}, UAP and GAP. To evaluate our method combined with the momentum method~\cite{dong2018boosting}, we further compare with the state-of-the-art method, AT-UAP which adopts the Adam optimizer.

\textbf{Hyper-parameters.} For fair comparison with previous works~\cite{moosavi2017universal,poursaeed2018generative,mopuri2018nag,zhang2020understanding,zhang2021data,li2022learning}, we set the maximum perturbation $\epsilon =10/255$. For SPGD, we follow the setting in ~\cite{shafahi2020universal} with the maximum cross-entropy loss $\beta =9$, and the step size $\alpha =1/255$. Moreover, the number of epochs for the SPGD and SGA is set to 20. To ensure the same number of perturbation iterations, the batch size of SPGD and the outer large batch size $|\boldsymbol{x}^{\mathrm{LB}}|$ in SGA are both set to 250. For SGA, the inner small batch size of $|\boldsymbol{x}^{\mathrm{SB}}|$ for AlexNet, GoogleNet, and the other three models is set to 1, 2, and 10 respectively, and the inner iteration number $M$ is set to twice the number of $|\boldsymbol{x}^{\mathrm{LB}}|/|\boldsymbol{x}^{\mathrm{SB}}|$ for AlexNet and GoogleNet, and four times in the remaining three models. In the momentum methods, the decay factor is set to 0.9.

\subsection{Generalization performance of UAP}
We first perform universal adversarial attacks under the white-box and black-box settings respectively and evaluate the overall performance of our proposed SGA with baselines on the ImageNet validation set.

\textbf{White-Box Setting.} The results on five models of various UAP generation methods in the white-box scenario are depicted in Table \ref{white-box generalization}. In addition to our method, the results of UAP, SV-UAP, NAG, GAP, DF-UAP, Cos-UAP, and AT-UAP are reported as in the original papers. We can observe that our method achieves the highest attack performance across all the models. For stronger models with deeper networks, the advantage of our proposed method is even clearer, \eg improvement of over 2\% in the ResNet152 model. The results demonstrate that the UAPs generated by our method can generalize better to unknown samples.

\textbf{Black-Box Setting.} We also evaluate the proposed method with regular comparison methods, \textit{i.e.}, UAP, GAP, and SPGD in the black-box scenario. Both SPGD and SGA methods implement two different loss functions, \textit{i.e.}, logit loss and cross-entropy loss, namely SPGD-logit, SPGD-cls, SGA-logit, and SGA-cls respectively. We craft UAPs on normally trained models and evaluate them on all the five networks. As shown in Table \ref{base}, the SGA can significantly improve the fooling ratio across all the models on both loss functions. For the UAPs crafted on the AlexNet model, the average fooling ratio increases from 54.96\% and 56.38 \% to 60.21\% and 61.71\%, respectively. Furthermore, the average fooling ratio achieved by our methods outperforms comparison UAP attack methods by 3.60\% $\sim$ 19.28\% on average, which verifies SGA method can effectively enhance the cross-model generalization ability of UAP. The visualization results are provided in the supplementary.

\begin{figure}[!ht]
	\centering
	\includegraphics[width=0.90\linewidth]{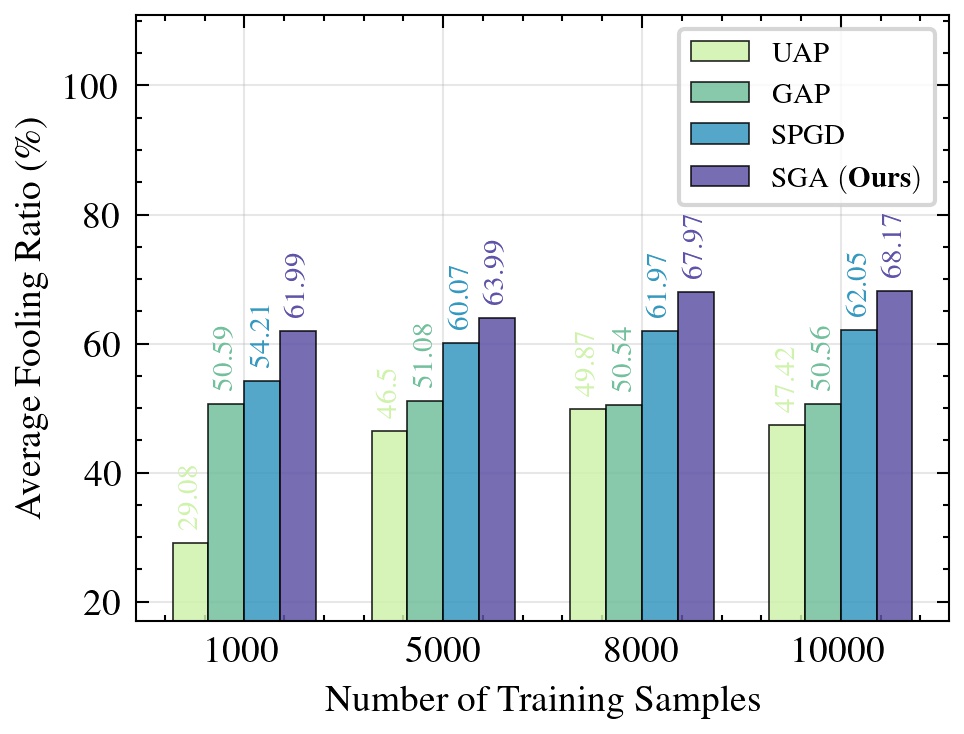}
	\caption{Average fooling ratio (\%) on five models versus the number of training samples. The UAPs are crafted by SPGD and SGA methods on the VGG16 model.}
	\label{img_num}
\end{figure}

\subsection{UAPs via different number of training samples}
We further investigate the generation of effective UAPs via different numbers of training samples. In the following experiments, without otherwise stated, the clipped cross-entropy loss is used in SPGD and SGA methods.

\textbf{Limit-Sample Setting.} Considering that the acquisition of a large number of training samples is difficult in some tasks, we randomly select 500 training samples from the ImageNet training set to simulate the limit-sample scenario. In Table \ref{limit_500}, the fooling ratio across five models achieved by our method remarkably outperforms other comparison methods, showing that SGA can fully exploit the potential of limited samples. It is worth mentioning that the proposed method reaches about 90\% fooling ratio on AlexNet, VGG16, and VGG19 models, which is close to the attack performance achieved on 10,000 training samples.

\begin{table*}[!ht]
	\begin{center}
		\renewcommand{\arraystretch}{1.0}
		\scalebox{0.85}{
			\begin{tabular}{|c|c|cccccc|}
				\hline
				Model                                            & Method & AlexNet         & GoogleNet       & VGG16           & VGG19           & ResNet152       & Average        \\ \hline \hline
				\multirow{5}{*}{AlexNet}                         & AT-UAP & 97.01*          & 47.31           & 62.37           & 57.72           & 33.40           & 59.56          \\
				& M-SPGD & 96.95*          & 43.29           & 62.26           & 56.61           & 31.91           & 58.20          \\
				& M-SGA (\textbf{Ours})  & \textbf{97.43*} & \textbf{49.71}  & \textbf{66.41}  & \textbf{60.96}  & \textbf{35.76}  & \textbf{62.05} \\ \cline{2-8} 
				& N-SPGD & 96.96*          & 39.87           & 60.27           & 56.00           & 29.90           & 56.60          \\
				& N-SGA (\textbf{Ours})  & \textbf{97.35*} & \textbf{47.93}  & \textbf{65.74}  & \textbf{60.17}  & \textbf{35.00}  & \textbf{61.24} \\ \hline  
				\multirow{5}{*}{GoogleNet}                       & AT-UAP & 55.90           & 90.82*          & 78.71           & 76.01           & 54.49           & 71.19          \\
				& M-SPGD & 53.06           & \textbf{92.20*} & 78.43           & 76.05           & 54.62           & 70.87          \\
				& M-SGA (\textbf{Ours})  & \textbf{62.56}  & 92.12*          & \textbf{83.62}  & \textbf{82.11}  & \textbf{59.09}  & \textbf{75.90} \\ \cline{2-8} 
				& N-SPGD & 53.55           & 92.23*          & 77.93           & 76.03           & 53.60           & 70.67          \\
				& N-SGA (\textbf{Ours})  & \textbf{63.43}  & \textbf{91.67*} & \textbf{84.23}  & \textbf{82.17}  & \textbf{58.41}  & \textbf{75.98} \\ \hline 
				\multirow{5}{*}{VGG16}                           & AT-UAP & 45.58           & 53.63           & 97.51*          & 91.53           & 47.16           & 67.08          \\
				& M-SPGD & 44.30           & 49.19           & 97.43*          & 91.06           & 43.51           & 65.10          \\
				& M-SGA (\textbf{Ours})  & \textbf{49.02}  & \textbf{55.78}  & \textbf{98.36*} & \textbf{94.17}  & \textbf{49.02}  & \textbf{69.27} \\ \cline{2-8} 
				& N-SPGD & 44.49           & 49.16           & 97.39*          & 90.55           & 42.45           & 64.81          \\
				& N-SGA (\textbf{Ours})  & \textbf{48.74}  & \textbf{55.61}  & \textbf{98.36*} & \textbf{93.98}  & \textbf{50.17}  & \textbf{69.37} \\ \hline 
				\multirow{5}{*}{VGG19}                           & AT-UAP & 46.04           & 52.58           & 93.49           & 97.56*          & 43.53           & 66.64          \\
				& M-SPGD & 44.89           & 45.26           & 91.25           & 96.36*          & 42.47           & 64.04          \\
				& M-SGA (\textbf{Ours})  & \textbf{50.67}  & \textbf{56.87}  & \textbf{95.52}  & \textbf{97.69*} & \textbf{51.08}  & \textbf{70.37} \\ \cline{2-8} 
				& N-SPGD & 45.27           & 52.18           & 93.48           & 96.82*          & 46.46           & 66.84          \\
				& N-SGA (\textbf{Ours})  & \textbf{50.77}  & \textbf{56.95}  & \textbf{95.58}  & \textbf{97.73*} & \textbf{51.78}  & \textbf{70.56} \\ \hline 
				\multicolumn{1}{|l|}{\multirow{5}{*}{ResNet152}} & AT-UAP & 47.33           & 61.32           & \textbf{81.93}  & 78.72           & 91.52*          & 72.16          \\
				\multicolumn{1}{|l|}{}                           & M-SPGD & 48.84           & 58.84           & 79.46           & 76.37           & 92.83*          & 71.27          \\
				\multicolumn{1}{|l|}{}                           & M-SGA (\textbf{Ours})  & \textbf{51.59}  & \textbf{64.05}  & 81.77           & \textbf{79.01}  & \textbf{94.04*} & \textbf{74.09} \\ \cline{2-8} 
				\multicolumn{1}{|l|}{}                           & N-SPGD & 49.00           & 58.01           & 78.65           & 76.49           & 92.87*          & 71.00          \\
				\multicolumn{1}{|l|}{}                           & N-SGA (\textbf{Ours})  & \textbf{51.50}  & \textbf{65.29}  & \textbf{83.02}  & \textbf{79.41}  & \textbf{93.92*} & \textbf{74.63} \\ \hline
			\end{tabular}
		}
	\end{center}
	\caption{The fooling ratio (\%) on five models in the single model setting by gradient-based UAP attack methods enhanced by Momentum and Nesterov accelerated gradient methods respectively. * indicates the white-box model.}
	\label{Momentum method}
\end{table*} 

\textbf{Diverse-Sample Setting.} Furthermore, we explore the effect of the number of training samples on the attack performance. The results are shown in Fig \ref{img_num}. Compared with the baseline attack, our method can promote the improvement of the generalization ability of UAP under different numbers of training samples. However, when the number of samples reaches a certain number, the attack performance does not increase anymore correspondingly. The phenomenon reflects the limitations of excessively increasing the number of training samples.

\subsection{Combination with transferability methods}
In adversarial attacks, Momentum~\cite{dong2018boosting} and Nesterov accelerated gradient~\cite{lin2019nesterov} are considered as the baselines to improve model transferability and can be well integrated with other methods. Therefore, we further incorporate SPGD and SGA with these two baseline methods, denoted as M-SPGD, N-SPGD, M-SGA, and N-SGA, under single-model and ensemble-model settings ~\cite{liu2016delving} respectively.

\textbf{Single-Model Setting.} In the integrated momentum methods, we report the results in the original paper of the current state-of-the-art attack method, AT-UAP. Experimental results on the ImageNet validation set are listed in Table \ref{Momentum method}. We can see that our method outperforms the baseline method by a large margin on all black-box models, while also improving the attack performance of white-box attacks. In general, the SGA-based method consistently outperforms the baseline method by 2.82\% $\sim$ 6.33\% and exceeds the current state-of-the-art method in attack performance.

\begin{table*}[!ht]
	\begin{center}
		\renewcommand{\arraystretch}{1.0}
		\scalebox{0.865}{
			\begin{tabular}{|c|c|ccccccc|}
				\hline
				Model                     & Method & AlexNet         & VGG16           & GoogleNet      & VGG19          & \multicolumn{1}{l}{ResNet50} & ResNet152      & Average        \\ \hline \hline
				\multirow{6}{*}{Ensemble} & SPGD   & 93.77*          & 94.28*          & 63.47          & 86.45          & 58.45                        & 48.39          & 74.10          \\
				& SGA (\textbf{Ours})    & \textbf{94.54*} & \textbf{95.50*} & \textbf{66.16} & \textbf{88.54} & \textbf{62.06}               & \textbf{51.04} & \textbf{76.31} \\ \cline{2-9} 
				& M-SPGD & 93.91*          & 95.21*          & 61.05          & 86.92          & 60.49                        & 48.72          & 74.38          \\
				& M-SGA (\textbf{Ours})  & \textbf{94.65*} & \textbf{96.03*} & \textbf{66.83} & \textbf{90.00} & \textbf{62.72}               & \textbf{52.39} & \textbf{77.10} \\ \cline{2-9} 
				& N-SPGD & 93.94*          & 95.31*          & 60.41          & 87.33          & 59.39                        & 47.99          & 74.06          \\
				& N-SGA (\textbf{Ours})  & \textbf{94.20*} & \textbf{96.19*} & \textbf{64.49} & \textbf{89.21} & \textbf{65.23}               & \textbf{53.36} & \textbf{77.11} \\ \hline
			\end{tabular}
		}
	\end{center}
	\caption{The fooling ratio (\%) on five models in the ensemble model setting by gradient-based UAP attack methods enhanced by Momentum and Nesterov accelerated gradient methods respectively. The UAPs are crafted on the ensemble models, \textit{i.e.}, AlexNet and VGG16.}
	\label{ensemble}
\end{table*} 

\begin{figure*}[ht!]
	\centering
	\includegraphics[width=0.90\linewidth]{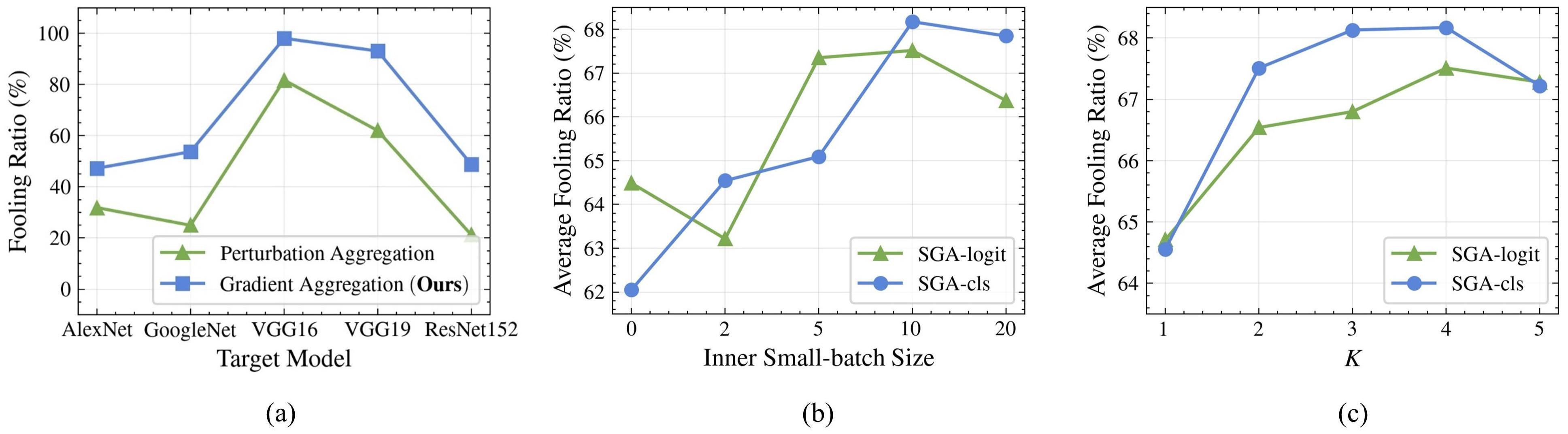}
	\caption{Ablation study on gradient aggregation, inner small-batch size and inner iteration number. (a) represents the average fooling ratio (\%) of five models using different types of aggregation methods, \textit{i.e.}, perturbation aggregation and gradient aggregation. (b) represents the average fooling ratio (\%) of five models with different inner batch sizes. (c) represents the average fooling ratio (\%) of five models with different inner iteration numbers. The UAPs are crafted by the VGG16 model under SGA.}
	\label{vgg16_aba}
\end{figure*}

\textbf{Ensemble-Model Setting.} Here we implement the model ensemble method in~\cite{dong2018boosting} which averages the loss functions of two normally trained models, \textit{i.e.}, AlexNet and VGG16. Except for the five widely used models, we also test the transferability on the ResNet50. The results are summarized in Table \ref{ensemble}. The SGA-based methods can greatly improve the attack performance of the baselines across all models, achieving the average fooling ratio from 74.1\%, 74.38\%, 74.06\% to 76.31\%, 77.10\% and 77.11\% respectively. Such compelling results verify the effectiveness of our method in combination with gradient optimization and model ensemble methods for improving transferability.

\subsection{Ablation study}
In this subsection, we conduct a series of ablation experiments to explore the impact of the gradient aggregation and the hyper-parameters for inner small-batch size and inner iteration number. All the UAPs are generated on VGG16 and evaluated on all the five widely used models. The results for the remaining models are provided in the supplementary.

\textbf{On the gradient aggregation.} To demonstrate that the gain of SGA is not by increasing the number of gradient calculations, we report the results of SGA with two types of aggregation methods, \textit{i.e.}, perturbation aggregation and gradient aggregation. Perturbation aggregation is to update $\delta$ by using gradients without accumulation as in SPGD, which can be easily achieved by utilizing the inner iterative perturbation $\delta _{M}^{\mathrm{inner}}$ to update UAP by $\delta \gets \delta _{M}^{\mathrm{inner}}$. The results are shown in Fig \ref{vgg16_aba} (a). The gradient aggregation method achieves remarkably better attack performance than the perturbation aggregation method, indicating the effectiveness of gradient aggregation to mitigate the gradient vanishing. 

\textbf{On the inner small-batch size.} We then investigate the impact of inner batch size $|\boldsymbol{x}^{\mathrm{SB}}|$ on the attack performance of SGA. We try different inner batch sizes of 0, 2, 5, 10, and 20 respectively. When the inner batch size is set to 0, SGA degrades to SPGD. As shown in Fig \ref{vgg16_aba} (b), the attack performance is excellent when $|\boldsymbol{x}^{\mathrm{SB}}|=10$. An appropriate batch can help enhance the generalization of UAP.

\textbf{On the inner iteration number.} After determining the inner batch size, we continue to explore the effect of the inner iteration number $M$. We set the inner iteration number as $M=K\times \left| \boldsymbol{x}^{\mathrm{LB}} \right|/\left| \boldsymbol{x}^{\mathrm{SB}} \right|$, where K represents the average number of times each image is traversed in one inner iteration. Fig \ref{vgg16_aba} (c) shows that with the increase of the iteration number, the attack performance reaches the peak when $K=4$. Similarly, an appropriate iteration number can help enhance the generalization of UAP. Besides, the discussion of the time consuming are provided in the supplementary.


\section{Conclusion}
In this paper, we propose a stochastic gradient aggregation (SGA) method to cope with the two main issues, \textit{i.e.}, gradient instability and quantization error in the existing UAP attack methods with mini-batch training. Specifically, we first perform multiple rounds of inner pre-search by implementing small-batch optimization. Then SGA aggregates all the gradients of the inner iterations as a one-step gradient estimate and updates the outer adversarial perturbations by using the aggregated gradients. Extensive experiments demonstrate that the proposed method can significantly enhance the generalization ability of the baseline methods in various settings and reach an average fooling ratio of 95.95\% exceeding the state-of-the-art methods.

\section{Acknowledgements}
This work was supported in part by the National Natural Science Foundation of China under Grant No.62276030 and 62236003, in part by BUPT Excellent Ph.D. Students Foundation No.CX2023111 and supported by Program for Youth Innovative Research Team of BUPT No.2023QNTD02.

{\small
\bibliographystyle{ieee_fullname}
\bibliography{egbib}

\begin{thebibliography}{10}\itemsep=-1pt

\bibitem{berger2022stereoscopic}
Zachary Berger, Parth Agrawal, Tian~Yu Liu, Stefano Soatto, and Alex Wong.
\newblock Stereoscopic universal perturbations across different architectures
  and datasets.
\newblock In {\em CVPR}, pages 15180--15190, 2022.

\bibitem{cheng2021fast}
Yaya Cheng, Jingkuan Song, Xiaosu Zhu, Qilong Zhang, Lianli Gao, and Heng~Tao
  Shen.
\newblock Fast gradient non-sign methods.
\newblock {\em arXiv preprint arXiv:2110.12734}, 2021.

\bibitem{co2021universal}
Kenneth~T Co, Luis Mu{\~n}oz-Gonz{\'a}lez, Leslie Kanthan, Ben Glocker, and
  Emil~C Lupu.
\newblock Universal adversarial robustness of texture and shape-biased models.
\newblock In {\em ICIP}, pages 799--803. IEEE, 2021.

\bibitem{ding2021beyond}
Wenjie Ding, Xing Wei, Rongrong Ji, Xiaopeng Hong, Qi Tian, and Yihong Gong.
\newblock Beyond universal person re-identification attack.
\newblock {\em IEEE transactions on information forensics and security},
  16:3442--3455, 2021.

\bibitem{dong2018boosting}
Yinpeng Dong, Fangzhou Liao, Tianyu Pang, Hang Su, Jun Zhu, Xiaolin Hu, and
  Jianguo Li.
\newblock Boosting adversarial attacks with momentum.
\newblock In {\em CVPR}, pages 9185--9193, 2018.

\bibitem{dong2019evading}
Yinpeng Dong, Tianyu Pang, Hang Su, and Jun Zhu.
\newblock Evading defenses to transferable adversarial examples by
  translation-invariant attacks.
\newblock In {\em CVPR}, pages 4312--4321, 2019.

\bibitem{goodfellow2014explaining}
Ian~J. Goodfellow, Jonathon Shlens, and Christian Szegedy.
\newblock Explaining and harnessing adversarial examples.
\newblock In {\em ICLR}, 2015.

\bibitem{he2016deep}
Kaiming He, Xiangyu Zhang, Shaoqing Ren, and Jian Sun.
\newblock Deep residual learning for image recognition.
\newblock In {\em CVPR}, pages 770--778, 2016.

\bibitem{hinton2012imagenet}
Geoffrey~E Hinton, Alex Krizhevsky, and Ilya Sutskever.
\newblock Imagenet classification with deep convolutional neural networks.
\newblock {\em NIPS}, 25(1106-1114):1, 2012.

\bibitem{hoffer2017train}
Elad Hoffer, Itay Hubara, and Daniel Soudry.
\newblock Train longer, generalize better: closing the generalization gap in
  large batch training of neural networks.
\newblock {\em NIPS}, 30, 2017.

\bibitem{howard2017mobilenets}
Andrew~G Howard, Menglong Zhu, Bo Chen, Dmitry Kalenichenko, Weijun Wang,
  Tobias Weyand, Marco Andreetto, and Hartwig Adam.
\newblock Mobilenets: Efficient convolutional neural networks for mobile vision
  applications.
\newblock {\em arXiv preprint arXiv:1704.04861}, 2017.

\bibitem{hu2018squeeze}
Jie Hu, Li Shen, and Gang Sun.
\newblock Squeeze-and-excitation networks.
\newblock In {\em CVPR}, pages 7132--7141, 2018.

\bibitem{keskar2016large}
Nitish~Shirish Keskar, Dheevatsa Mudigere, Jorge Nocedal, Mikhail Smelyanskiy,
  and Ping Tak~Peter Tang.
\newblock On large-batch training for deep learning: Generalization gap and
  sharp minima.
\newblock In {\em ICLR}, 2017.

\bibitem{khrulkov2018art}
Valentin Khrulkov and Ivan Oseledets.
\newblock Art of singular vectors and universal adversarial perturbations.
\newblock In {\em CVPR}, pages 8562--8570, 2018.

\bibitem{kurakin2018adversarial}
Alexey Kurakin, Ian~J Goodfellow, and Samy Bengio.
\newblock Adversarial examples in the physical world.
\newblock In {\em Artificial intelligence safety and security}, pages 99--112.
  Chapman and Hall/CRC, 2018.

\bibitem{li2022learning}
Maosen Li, Yanhua Yang, Kun Wei, Xu Yang, and Heng Huang.
\newblock Learning universal adversarial perturbation by adversarial example.
\newblock In {\em AAAI}, 2022.

\bibitem{lin2019nesterov}
Jiadong Lin, Chuanbiao Song, Kun He, Liwei Wang, and John~E. Hopcroft.
\newblock Nesterov accelerated gradient and scale invariance for adversarial
  attacks.
\newblock In {\em ICLR}, 2020.

\bibitem{liu2016delving}
Yanpei Liu, Xinyun Chen, Chang Liu, and Dawn Song.
\newblock Delving into transferable adversarial examples and black-box attacks.
\newblock In {\em ICLR}, 2017.

\bibitem{madry2017towards}
Aleksander Madry, Aleksandar Makelov, Ludwig Schmidt, Dimitris Tsipras, and
  Adrian Vladu.
\newblock Towards deep learning models resistant to adversarial attacks.
\newblock In {\em ICLR}, 2018.

\bibitem{masters2018revisiting}
Dominic Masters and Carlo Luschi.
\newblock Revisiting small batch training for deep neural networks.
\newblock {\em arXiv preprint arXiv:1804.07612}, 2018.

\bibitem{matachana2020robustness}
Alberto~G Matachana, Kenneth~T Co, Luis Mu{\~n}oz-Gonz{\'a}lez, David Martinez,
  and Emil~C Lupu.
\newblock Robustness and transferability of universal attacks on compressed
  models.
\newblock {\em arXiv preprint arXiv:2012.06024}, 2020.

\bibitem{moosavi2017universal}
Seyed-Mohsen Moosavi-Dezfooli, Alhussein Fawzi, Omar Fawzi, and Pascal
  Frossard.
\newblock Universal adversarial perturbations.
\newblock In {\em CVPR}, pages 1765--1773, 2017.

\bibitem{moosavi2016deepfool}
Seyed-Mohsen Moosavi-Dezfooli, Alhussein Fawzi, and Pascal Frossard.
\newblock Deepfool: a simple and accurate method to fool deep neural networks.
\newblock In {\em CVPR}, pages 2574--2582, 2016.

\bibitem{mopuri2018nag}
Konda~Reddy Mopuri, Utkarsh Ojha, Utsav Garg, and R~Venkatesh Babu.
\newblock Nag: Network for adversary generation.
\newblock In {\em CVPR}, pages 742--751, 2018.

\bibitem{papernot2017practical}
Nicolas Papernot, Patrick McDaniel, Ian Goodfellow, Somesh Jha, Z~Berkay Celik,
  and Ananthram Swami.
\newblock Practical black-box attacks against machine learning.
\newblock In {\em Proceedings of the 2017 ACM on Asia conference on computer
  and communications security}, pages 506--519, 2017.

\bibitem{papernot2016limitations}
Nicolas Papernot, Patrick McDaniel, Somesh Jha, Matt Fredrikson, Z~Berkay
  Celik, and Ananthram Swami.
\newblock The limitations of deep learning in adversarial settings.
\newblock In {\em 2016 IEEE European symposium on security and privacy
  (EuroS\&P)}, pages 372--387. IEEE, 2016.

\bibitem{peng2022fingerprinting}
Zirui Peng, Shaofeng Li, Guoxing Chen, Cheng Zhang, Haojin Zhu, and Minhui Xue.
\newblock Fingerprinting deep neural networks globally via universal
  adversarial perturbations.
\newblock In {\em CVPR}, pages 13430--13439, 2022.

\bibitem{poursaeed2018generative}
Omid Poursaeed, Isay Katsman, Bicheng Gao, and Serge Belongie.
\newblock Generative adversarial perturbations.
\newblock In {\em CVPR}, pages 4422--4431, 2018.

\bibitem{qin2022boosting}
Zeyu Qin, Yanbo Fan, Yi Liu, Li Shen, Yong Zhang, Jue Wang, and Baoyuan Wu.
\newblock Boosting the transferability of adversarial attacks with reverse
  adversarial perturbation.
\newblock {\em arXiv preprint arXiv:2210.05968}, 2022.

\bibitem{rampini2021universal}
Arianna Rampini, Franco Pestarini, Luca Cosmo, Simone Melzi, and Emanuele
  Rodola.
\newblock Universal spectral adversarial attacks for deformable shapes.
\newblock In {\em CVPR}, pages 3216--3226, 2021.

\bibitem{russakovsky2015imagenet}
Olga Russakovsky, Jia Deng, Hao Su, Jonathan Krause, Sanjeev Satheesh, Sean Ma,
  Zhiheng Huang, Andrej Karpathy, Aditya Khosla, Michael Bernstein, et~al.
\newblock Imagenet large scale visual recognition challenge.
\newblock {\em International journal of computer vision}, 115(3):211--252,
  2015.

\bibitem{shafahi2019adversarial}
Ali Shafahi, Mahyar Najibi, Mohammad~Amin Ghiasi, Zheng Xu, John Dickerson,
  Christoph Studer, Larry~S Davis, Gavin Taylor, and Tom Goldstein.
\newblock Adversarial training for free!
\newblock {\em NIPS}, 32, 2019.

\bibitem{shafahi2020universal}
Ali Shafahi, Mahyar Najibi, Zheng Xu, John Dickerson, Larry~S Davis, and Tom
  Goldstein.
\newblock Universal adversarial training.
\newblock In {\em AAAI}, volume~34, pages 5636--5643, 2020.

\bibitem{simonyan2014very}
Karen Simonyan and Andrew Zisserman.
\newblock Very deep convolutional networks for large-scale image recognition.
\newblock In {\em ICLR}, 2015.

\bibitem{szegedy2015going}
Christian Szegedy, Wei Liu, Yangqing Jia, Pierre Sermanet, Scott Reed, Dragomir
  Anguelov, Dumitru Erhan, Vincent Vanhoucke, and Andrew Rabinovich.
\newblock Going deeper with convolutions.
\newblock In {\em CVPR}, pages 1--9, 2015.

\bibitem{szegedy2013intriguing}
Christian Szegedy, Wojciech Zaremba, Ilya Sutskever, Joan Bruna, Dumitru Erhan,
  Ian~J. Goodfellow, and Rob Fergus.
\newblock Intriguing properties of neural networks.
\newblock In {\em ICLR}, 2014.

\bibitem{wang2022boosting}
Jiafeng Wang, Zhaoyu Chen, Kaixun Jiang, Dingkang Yang, Lingyi Hong, Yan Wang,
  and Wenqiang Zhang.
\newblock Boosting the transferability of adversarial attacks with global
  momentum initialization.
\newblock {\em arXiv preprint arXiv:2211.11236}, 2022.

\bibitem{wang2021enhancing}
Xiaosen Wang and Kun He.
\newblock Enhancing the transferability of adversarial attacks through variance
  tuning.
\newblock In {\em CVPR}, pages 1924--1933, 2021.

\bibitem{xiao2018generating}
Chaowei Xiao, Bo Li, Jun{-}Yan Zhu, Warren He, Mingyan Liu, and Dawn Song.
\newblock Generating adversarial examples with adversarial networks.
\newblock In {\em IJCAI}, 2018.

\bibitem{xie2019improving}
Cihang Xie, Zhishuai Zhang, Yuyin Zhou, Song Bai, Jianyu Wang, Zhou Ren, and
  Alan~L Yuille.
\newblock Improving transferability of adversarial examples with input
  diversity.
\newblock In {\em CVPR}, pages 2730--2739, 2019.

\bibitem{xie2021enabling}
Yi Xie, Zhuohang Li, Cong Shi, Jian Liu, Yingying Chen, and Bo Yuan.
\newblock Enabling fast and universal audio adversarial attack using generative
  model.
\newblock In {\em AAAI}, volume~35, pages 14129--14137, 2021.

\bibitem{xiong2022stochastic}
Yifeng Xiong, Jiadong Lin, Min Zhang, John~E Hopcroft, and Kun He.
\newblock Stochastic variance reduced ensemble adversarial attack for boosting
  the adversarial transferability.
\newblock In {\em CVPR}, pages 14983--14992, 2022.

\bibitem{zhang2020cd}
Chaoning Zhang, Philipp Benz, Tooba Imtiaz, and In-So Kweon.
\newblock Cd-uap: Class discriminative universal adversarial perturbation.
\newblock In {\em AAAI}, volume~34, pages 6754--6761, 2020.

\bibitem{zhang2020understanding}
Chaoning Zhang, Philipp Benz, Tooba Imtiaz, and In~So Kweon.
\newblock Understanding adversarial examples from the mutual influence of
  images and perturbations.
\newblock In {\em CVPR}, pages 14521--14530, 2020.

\bibitem{zhang2021data}
Chaoning Zhang, Philipp Benz, Adil Karjauv, and In~So Kweon.
\newblock Data-free universal adversarial perturbation and black-box attack.
\newblock In {\em ICCV}, pages 7868--7877, 2021.

\bibitem{zhang2023unsupervised}
Yuhang Zhang, Weihong Deng, and Liang Zheng.
\newblock Unsupervised evaluation of out-of-distribution detection: A
  data-centric perspective.
\newblock {\em arXiv preprint arXiv:2302.08287}, 2023.

\bibitem{zhang2021relative}
Yuhang Zhang, Chengrui Wang, and Weihong Deng.
\newblock Relative uncertainty learning for facial expression recognition.
\newblock {\em NIPS}, 34:17616--17627, 2021.

\bibitem{zhang2022revisiting}
Yihua Zhang, Guanhua Zhang, Prashant Khanduri, Mingyi Hong, Shiyu Chang, and
  Sijia Liu.
\newblock Revisiting and advancing fast adversarial training through the lens
  of bi-level optimization.
\newblock In {\em ICML}, pages 26693--26712. PMLR, 2022.

\bibitem{zhao2021success}
Zhengyu Zhao, Zhuoran Liu, and Martha Larson.
\newblock On success and simplicity: A second look at transferable targeted
  attacks.
\newblock {\em NIPS}, 34:6115--6128, 2021.

\bibitem{zhong2020towards}
Yaoyao Zhong and Weihong Deng.
\newblock Towards transferable adversarial attack against deep face
  recognition.
\newblock {\em IEEE Transactions on Information Forensics and Security},
  16:1452--1466, 2020.

\bibitem{zhong2022opom}
Yaoyao Zhong and Weihong Deng.
\newblock Opom: Customized invisible cloak towards face privacy protection.
\newblock {\em IEEE Transactions on Pattern Analysis and Machine Intelligence},
  2022.

\end{thebibliography}
}

\end{document}